\documentclass{article} 
\usepackage{MLDD_workshop_2022,times}
\usepackage{booktabs}

\usepackage{amsmath,amsfonts,bm}

\def\eqref#1{equation~\ref{#1}}

\def\1{\bm{1}}

\def\evd{{d}}

\def\evr{{r}}

\def\evx{{x}}
\def\evy{{y}}

\def\mF{{\bm{F}}}

\DeclareMathAlphabet{\mathsfit}{\encodingdefault}{\sfdefault}{m}{sl}
\SetMathAlphabet{\mathsfit}{bold}{\encodingdefault}{\sfdefault}{bx}{n}

\def\sX{{\mathbb{X}}}
\def\sY{{\mathbb{Y}}}

\def\emF{{F}}

\newcommand{\R}{\mathbb{R}}

\usepackage{hyperref}
\usepackage{url}

\usepackage{graphicx}
\usepackage{caption}
\usepackage{subcaption}

\title{SystemMatch: optimizing preclinical drug models to human clinical outcomes via generative latent-space matching}

\author{
Scott Gigante, Varsha G. Raghavan, Amanda M. Robinson,\\
\textbf{Robert A. Barton, Adeeb H. Rahman, Drausin F. Wulsin,}\\
\textbf{Jacques Banchereau, Noam Solomon, \& Luis F. Voloch} \\
Immunai\\
New York, NY 10016, USA
\AND
Fabian J. Theis \\
Helmholtz Munich and Technische Universit\"{a}t M\"{u}nchen \\
Germany \\
\texttt{fabian.theis@helmholtz-muenchen.de}
}

\iclrfinalcopy 
\begin{document}

\maketitle

\begin{abstract}    
Translating the relevance of preclinical models ($\textit{in vitro}$, animal models, or organoids) to their relevance in humans presents an important challenge during drug development. The rising abundance of single-cell genomic data from human tumors and tissue offers a new opportunity to optimize model systems by their similarity to targeted human cell types in disease. In this work, we introduce SystemMatch to assess the fit of preclinical model systems to an $\textit{in sapiens}$ target population and to recommend experimental changes to further optimize these systems. We demonstrate this through an application to developing $\textit{in vitro}$ systems to model human tumor-derived suppressive macrophages. We show with held-out $\textit{in vivo}$ controls that our pipeline successfully ranks macrophage subpopulations by their biological similarity to the target population, and apply this analysis to rank a series of 18 $\textit{in vitro}$ macrophage systems perturbed with a variety of cytokine stimulations. We extend this analysis to predict the behavior of 66 $\textit{in silico}$ model systems generated using a perturbational autoencoder and apply a $k$-medoids approach to recommend a subset of these model systems for further experimental development in order to fully explore the space of possible perturbations. Through this use case, we demonstrate a novel approach to model system development to generate a system more similar to human biology.
%
\end{abstract}

\section{Introduction}

Among most therapeutic areas, failures in clinical trials are common and costly. Failure rates of drugs entering Phase I trials have hit 90\% across most therapeutic areas \citep{mullard2016parsing}. Oncology in particular has one of the highest rates of clinical trial failures, with only 4\% \citep{mullard2016parsing} of therapies entering phase 1 FDA clinical trials ultimately being approved, despite having the most active clinical trials, with approximately 32\% of Phase I FDA clinical trials being in oncology \citep{thomas2016clinical}.
In part due to the desire to accelerate new medicines into the clinic to address unmet medical need, but also driven by competition in the industry, drug development organizations do not devote adequate time and resources to improve preclinical model systems that might be more predictive of clinical results \citep{honkala2021harnessing}.  Instead, the standard approach adopted by most of the industry relies on \textit{in vitro} and \textit{in vivo} tumor model systems that are poorly predictive of activity in patients due to the reductionist nature of the systems and inadequate attention devoted to understanding the molecular similarities and differences between preclinical and clinical data.

In this work, we describe an end-to-end machine learning pipeline, SystemMatch, that optimizes a preclinical model to best approximate the behavior of a target \textit{in sapiens} population to enable drug developers to quickly optimize preclinical models and identify those with the greatest predictive power for therapeutic priorities. SystemMatch uses single-cell genomic data from many preclinical models and evaluates them against a multi-study atlas of single-cell genomics data from the tumor or tissue of interest, helping decide which of these models is most likely to provide clinically meaningful predictions.
Furthermore, SystemMatch utilizes a Compositional Perturbational Autoencoder (CPA) \citep{lotfollahi2021compositional} to predict the behavior of single cells in previously untested combinations of experimental conditions. We use these predictions to recommend experimental changes to the preclinical models to enhance their similarity to the target population.

We demonstrate SystemMatch on a large multi-condition perturbational dataset of \textit{in vitro} differentiating macrophages at the single-cell level, and we compare these model systems' proximity to a target population of human tumor suppressive macrophages obtained from a multi-study single-cell macrophage atlas we collated, integrated, and annotated for this purpose. We demonstrate that SystemMatch produces systems that contradict the standard dogma for generating \textit{in vitro} suppressive macrophages, and we recommend further optimizations of our model system to generate systems with more predictive power for this purpose. This is, to our knowledge, the first computational pipeline for assessing and optimizing the fitness of preclinical models to an \textit{in sapiens} target population.

\section{Background}

\subsection{Single-cell omics}

Single cell omics refers to the quantitative characterization of biological phenotype at the cellular level. Early work in single-cell omics focused on the development of single-cell RNA sequencing technology \citep{kolodziejczyk2015technology}, in which the quantity of gene transcripts (or mRNA) in each cell is counted through complex microfluidic assays. Further work has expanded this technology to the measurement of many modalities, including genes, proteins, metabolites, transcripts, lipids and more \citep{chen2020single}, as well as multimodal data such as CITE-seq, which measures both RNA expression and protein abundance in the same assay \citep{citeseq}. As the quality and quantity of single-cell omics data rise, it is increasingly straightforward to precisely define rare cell types and hitherto poorly understood cellular heterogeneity (see, e.g., \citet{jaitin2014massively,zeisel2015cell}). Bulk RNA sequencing, on the other hand, refers to the sequencing of transcripts present in a large number of cells at the resolution of a cell type or tissue. Most large bulk datasets are not cell type specific, which makes it impractical to understand the phenotypic profile of specific cell subpopulations. For example, with bulk RNAseq data obtained from solid tumors, it is typically not possible to accurately understand the transcriptomic profile of specific immune subpopulations, like CD8 memory T-cells or immunosuppressive macrophages, despite recent work to deconvolve bulk data to cell type resolution \citep{newman2015robust,finotello2019molecular}.


\subsection{Preclinical model systems}

Due to both the cost and ethical implications of testing novel drugs in humans, most or all drugs are first tested in preclinical models, which range from microorganisms, to cell- and tissue-based models, to animal models including mice and non-human primates. Preclinical data are typically required for FDA approval to test a drug in humans~\citep{mcelvany2009fda} and are additionally used to prioritize selection of drugs to advance to clinical trial \citep{denayer2014animal}.  However, failures to translate success in preclinical models to humans have cast doubt on the predictive power of these models, prompting some to question their utility in drug prioritization (see, e.g., \citet{Schnabel2008,Suckling2008}), with some even going as far as to recommend forgoing animal models altogether and testing drugs directly in humans \citep{shanks2009animal}. On the other hand, reverse translational approaches seek to inform the development of preclinical models through the study of clinical success, creating an iterative process between preclinical and clinical studies to optimize later generations of drugs, the success of which can be seen, for example, in the development of EGFR tyrosine kinase inhibitors for the treatment of non-small cell lung carcinoma \citep{honkala2021harnessing}.

\subsection{Perturbation prediction}

Predicting cellular responses to perturbations is an important goal in computational biology.  \citet{ji2021machine} detail several uses for modeling perturbational single-cell data, including perturbation response prediction, target and mechanism prediction, perturbation interaction prediction, and chemical property prediction. Here, we introduce a new use case for perturbational modeling, which is to predict perturbations that will generate an optimal model system.  We generate \textit{in silico} predictions for a wide variety of possible perturbations, then select those closest to our target model system.  In this way, we can more rapidly converge on an ideal model system.

\section{Related work}

\subsection{Preclinical model system development}

Classical protocols of preclinical model system design use direct measurement of the preclinical system's phenotype to evaluate the quality of the system. For example, \citet{mia2014optimized} select among a set of possible protocols to generate \textit{in vitro} immunosuppressive macrophages by measuring a) secreted proteins known to be markers of immunosuppression and b) \textit{in vitro} suppression of T cells, finding that \textit{M-CSF} + \textit{IL-4} + \textit{IL-10} + \textit{TGF$\beta$} generates the most suppressive macrophages. \citet{fogg2020ovarian} take a different approach, culturing monocytes with ovarian cancer cell lines in order to understand the pathways activated by the cancer cells, finding an alternative pathway to macrophage polarization through \textit{TGF$\alpha$}. Reverse translational medicine, on the other hand, uses deep characterization of clinical response to existing drugs to understand the mechanisms of action of these drugs in order to design preclinical models that replicate the drug resistance mechanisms in humans \citep{honkala2021harnessing}. However, to our knowledge, no prior work has been done to optimize a preclinical model's phenotype in the high-dimensional space made visible through single-cell genomics.


\subsection{Perturbation response prediction}
Machine learning models have been used in a number of different ways to predict cellular perturbation response \cite{ji2021machine}.  A common setting is \emph{causal imputation} \cite{squires2020causal}, where a model is required to predict a response to an intervention in a particular context after having been trained on related interventions and contexts.  For example, \citet{squires2020causal} predict genomic response data by training on perturbations in some cell types and predicting response in other cell types.  \citet{lotfollahi2019scgen} introduces \textit{scGen} for causal imputation on single-cell gene expression data, which uses a variational autoencoder to represent interventions in latent space and then adds the interventions in latent space to an unperturbed representation to obtain the perturbational response.  In our setting, where we wish to predict perturbations that generate gene expression corresponding to an \textit{in vivo} model system, combinations of treatments are also important to consider.  Recently, the Compositional Perturbational Autoencoder \cite{lotfollahi2021compositional} was demonstrated to be able to predict combinations of gene knockouts from being trained on each knockout individually.  

\section{SystemMatch}

\begin{figure}[tb]
\centering
\includegraphics[width=\textwidth]{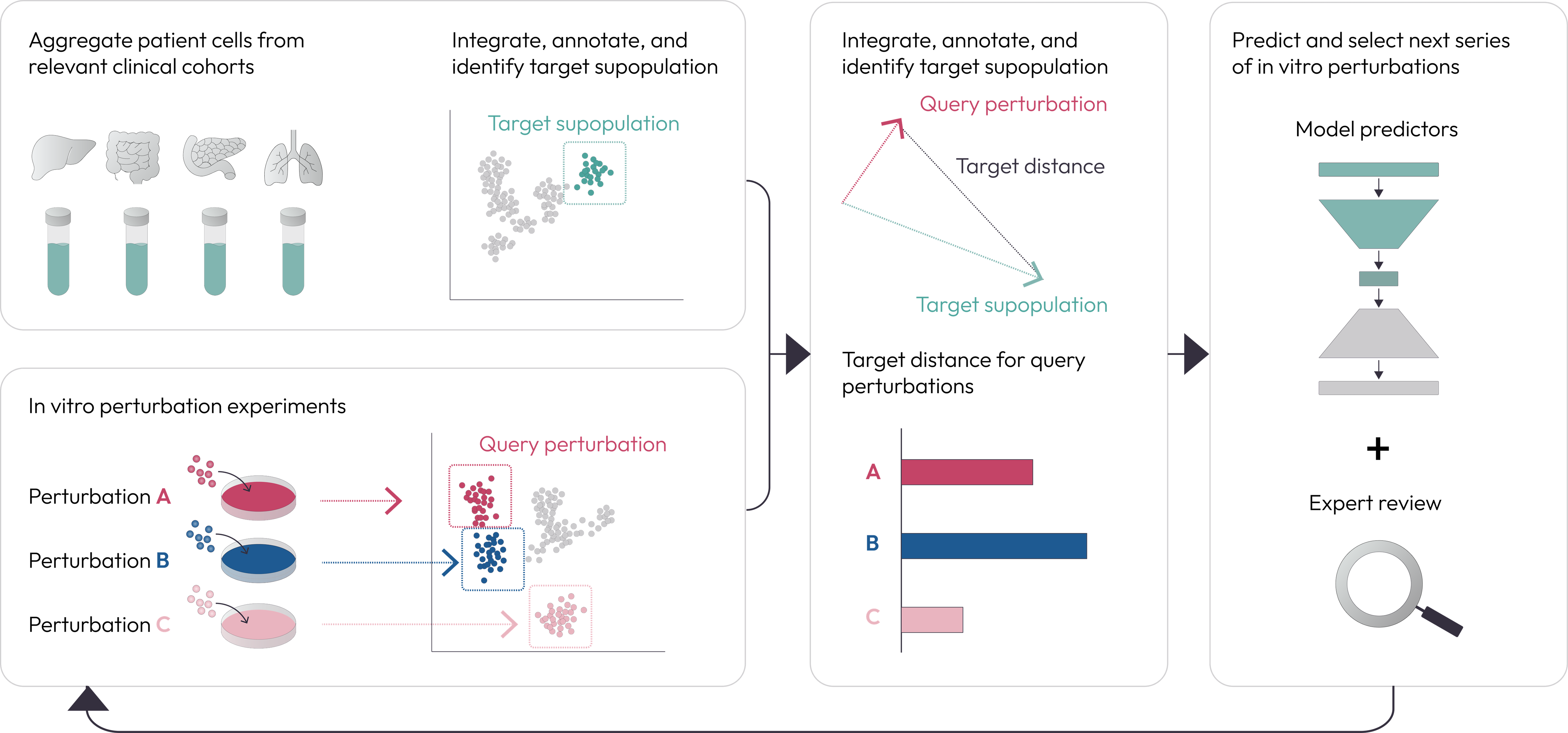}
\caption{Schematic of the iterative SystemMatch pipeline.}
\label{fig:schematic}
\end{figure}

Here, we introduce our iterative preclinical model optimization pipeline, SystemMatch. Fig.~\ref{fig:schematic} shows a schematic representation of the SystemMatch process. First, in order to define the \textit{target} population, we collect an atlas of single-cell data from relevant clinical cohorts giving a robust universal representation of the cell type of interest across multiple disease states. We apply a simple reference-based cell type annotation (see, e.g., \citet{seurat,scvi}) to only retain the cells associated with the target population (e.g., all macrophages). Next, we integrate this single-cell atlas using a fixed gene list derived from expert domain knowledge to capture only the biology relevant to the system at hand and use this as input to a batch correction algorithm (see, e.g., \citet{harmony,seurat,haghverdi2018mnn}). Within this integrated single-cell cell type--specific atlas, we annotate subtypes using a combination of graph-based clustering \citep{leiden} and expert domain knowledge. We choose the target population from these subtypes as the subtype that most closely recapitulates the phenotype of the target population as described in the literature (e.g., macrophages that express known markers of immunosuppression).

Second, we collect the \textit{query} model systems. We generate this data by developing many different model systems in parallel experiments (here, \textit{in vitro} macrophage differentiation and polarization) with multiple experimental conditions (here, stimulation with different cytokines and combinations thereof) selected from prior domain knowledge to generate a heterogeneous set of model systems. We perform single-cell sequencing on the cells of interest from each model system and then perform quality control on these datasets to remove low quality and outlier cells in order to obtain a clean single-cell resolution representation of each model system, and project these datasets to a subspace of the full gene expression matrix by retaining only those genes known to be related to the biological function of interest.

Next, we compute the distances in this subspace between each query dataset and the target population using some distance metric $m$. In the simplest case, $m$ is simply the L2 distance between the average (or "pseudobulked") gene expression of the query $\sX := \{\evx_i \in \R^d\}$ and the target $\sY := \{\evy_i \in \R^d\}$

$$m_{L2}(\sX, \sY) := \left\Vert \frac{\sum_i^{n_\sX} \evx_i}{n_\sX} - \frac{\sum_i^{n_\sY} \evy_i}{n_\sY} \right\Vert_2$$

where $n_\sX:=|X|$ and $n_\sY:=|Y|$. However, other more complex distance metrics between single-cell datasets could also be used, e.g. the Earth Mover's Distance (EMD) or Wasserstein metric \citep{kantorovich1960mathematical}

$$m_{EMD}(\sX, \sY) := \min\limits_\mF {\sum_{i=1}^{n_\sY}\sum_{j=1}^{n_\sX} \emF_{i,j}d(x_i,x_j)}$$

where $\mF$ is the $n_\sX \times n_\sY$ flow matrix with $\emF_{i,j} \ge 0$ and $d$ is a distance metric between cells, e.g. the Euclidean distance. Then, using our choice of $m$, we rank all query datasets to produce an ordered set of queries, with the query least distant from the target denoted the most representative model system of those tested.

However, this still leaves an open question: can we further improve the models beyond the original set of tested systems? To assist with this question, we employ generative deep neural networks to generate all possible combinations of conditions tested in the experimental queries to generate combinatorially many \textit{in silico} model systems. Then, in order to avoid over-reliance on the accuracy of these predictions, we leverage the \textit{in silico} queries to generate a search space of possible experimental conditions that are substantially different from the systems already tested. We use a modified form of the $k$-medoids algorithm \citep{kaufman2009finding} applied to the pseudo-bulked \textit{in silico} queries. Briefly, $k$-medoids selects $k$ equidistant model centroids as in $k$-means but requires that these centroids be selected from the existing data points. We extend this algorithm to enforce that all selected medoids are also equidistant from all existing experimentally tested queries, which leaves us with the smallest possible subset of experiments to run in the next iteration while ensuring that we do not leave any region of the search space untested. We can further combine these recommendations with expert knowledge by examining the genes driving the difference between our best-ranked query and the target population to propose additional experimental conditions.

Finally, we use the outputs of this pipeline to re-run the model system generation experiment, and we iterate upon this process until the generated model system is either sufficiently similar to the target population, or further iterations fail to improve upon the existing queries.

\begin{figure}[tb]
     \centering
     \begin{subfigure}[t]{0.35\textwidth}
         \vskip 0pt
         \centering
         \includegraphics[width=\textwidth]{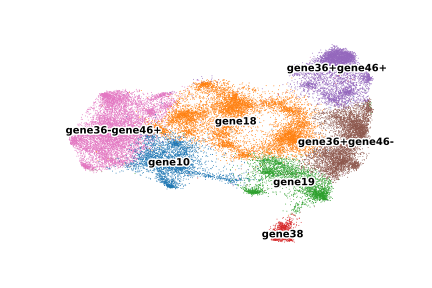} \\
         (a) \\
         \includegraphics[width=\textwidth]{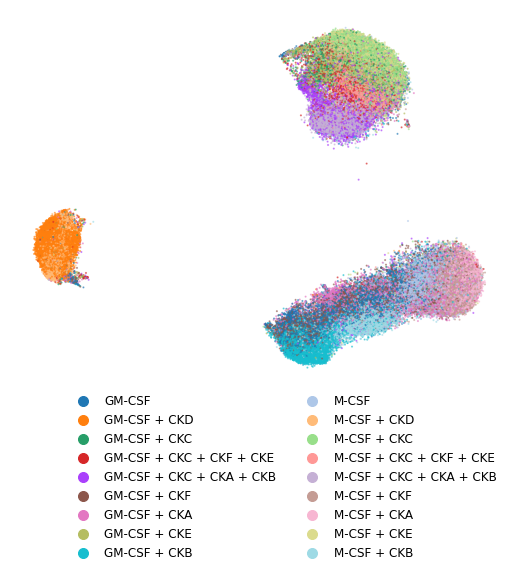} \\
         (b)
     \end{subfigure}
     \hfill
     \begin{subfigure}[t]{0.17\textwidth}
         \vskip 0pt
         \centering
         \includegraphics[width=\textwidth]{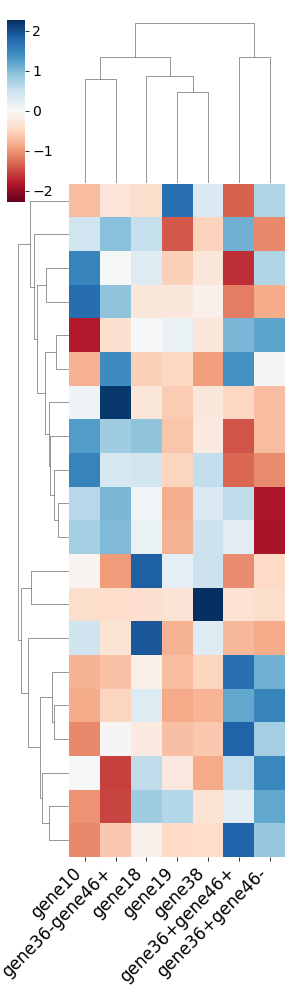}
     \end{subfigure}
     \hspace*{-0.9cm}
     \begin{subfigure}[t]{0.3925\textwidth}
         \vskip 0pt
         \centering
         \includegraphics[width=\textwidth]{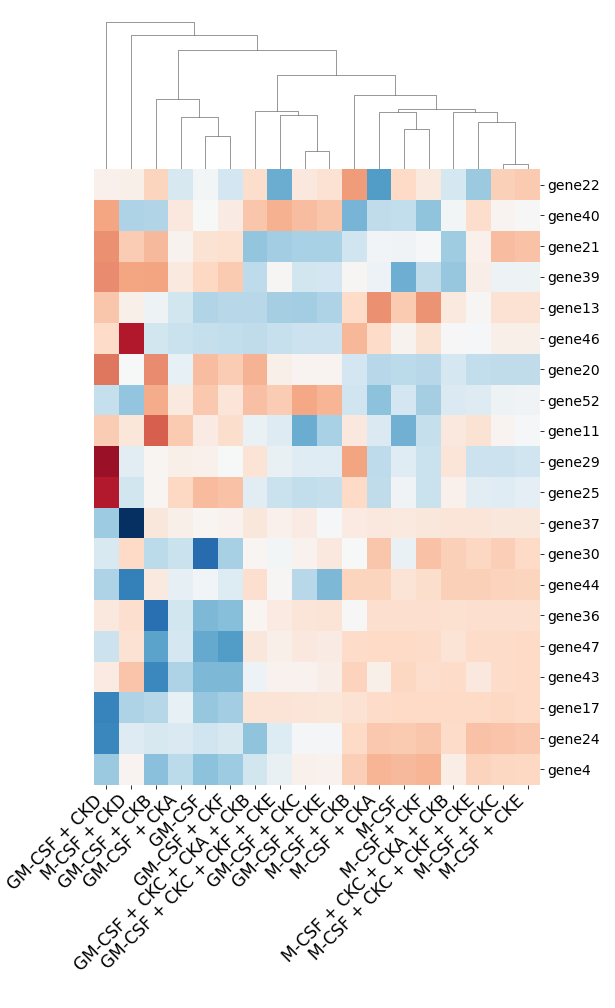}
     \end{subfigure} \\
     \vspace*{-1em}
     \hfill
     \begin{minipage}{0.57\textwidth}
     \centering
     (c)
     \end{minipage}
    \caption{Characterization of human tumor macrophage atlas and \textit{in vitro} stimulated macrophages. a) Human tumor macrophage atlas UMAP shows seven major clusters; b) \textit{in vitro} macrophage UMAP shows \textit{in vitro} conditions separate into three major groups; c) heatmap of major macrophage function genes in human macrophages (left) and \textit{in vitro} macrophages (right) shows the
    \textit{gene36-gene46+} cluster is most strongly associated with known suppressive response genes \textit{gene20} and \textit{gene46}, while this combination is absent in vitro.}
    \label{fig:dataset}
\end{figure}

\subsection{Generating \protect\textit{in silico} queries}

Since the space of possible model systems is extremely large and diverse, we are unable to test all possible experimental conditions \textit{in vitro}. Hence, to better explore the space of untested systems that could best emulate the target population, we predict the gene expression of combinations of the cytokine stimulations in our \textit{in vitro} experiment. We use CPA \cite{lotfollahi2021compositional} to generate \textit{in silico} perturbed cells for 66 additional conditions. Briefly, CPA is an autoencoder trained to decompose the data into disentangled latent representations for three key attributes of each cell: the cell's "basal state", the perturbation effect, and the covariate effect. These latent representations are combined in the decoder and trained via a reconstruction loss. To enforce independence of the latent representations, CPA is trained using a discriminator network and adversarial loss such that no signal from the observed perturbation or covariates is captured in the cellular basal state embedding. As the latent representations are combined by summation, CPA can combine multiple perturbations, allowing us to generate \textit{in silico} samples for combinations of cytokine stimulations. We further modify CPA to allow the incorporation of multiple covariates in order to account for experimental batch covariates included in our \textit{in vitro} dataset.

\subsection{Selecting new queries for experimental iteration}

To select a subset of the generated \textit{in silico} perturbations for experimental validation, we devise a scheme based on $k$-medoids \citep{kaufman2009finding} which ensures that the conditions selected for experimental validation are a) sufficiently different from the conditions already tested; and b) sufficiently heterogeneous to cover the space of possible perturbations. Briefly, we run the regular k-medoids algorithm, but at the medoid-update step of the algorithm, we consider each of the existing \textit{in vitro} perturbations as fixed medoids; this way, the $k$-medoids algorithm selects $k$ new perturbations for testing which are both heterogeneous and far from any of the previously tested conditions.

\section{Results}

\subsection{Characterizing \protect\textit{in vitro} and \protect\textit{in vivo} tumor associated macrophages}

We collated public and proprietary data from tumour-infiltrating immune cells from one study we generated internally and four public studies~\citep{zhang2021single,bassez2021single,yost2019clonal,qian2020pan}, comprising of 125 patients across five tumor types. We filtered these data to just retain the macrophage populations using reference-based mapping \citep{seurat}. We then filtered low-quality cells (defined as having fewer than 1500 genes with non-zero counts) and run batch integration across all 125 patients using Harmony~\citep{harmony} in a curated subspace of 319 macrophage function genes. We then ran Leiden clustering~\citep{leiden} and characterized each population independently according to expression of important marker genes. Finally, we selected the \textit{gene36-gene46+} population as our target population, as it expressed known markers associated with suppressive macrophages and correlated with poor clinical response to common immunotherapies.

\begin{figure}[tb]
     \centering
     \includegraphics[width=0.75\textwidth]{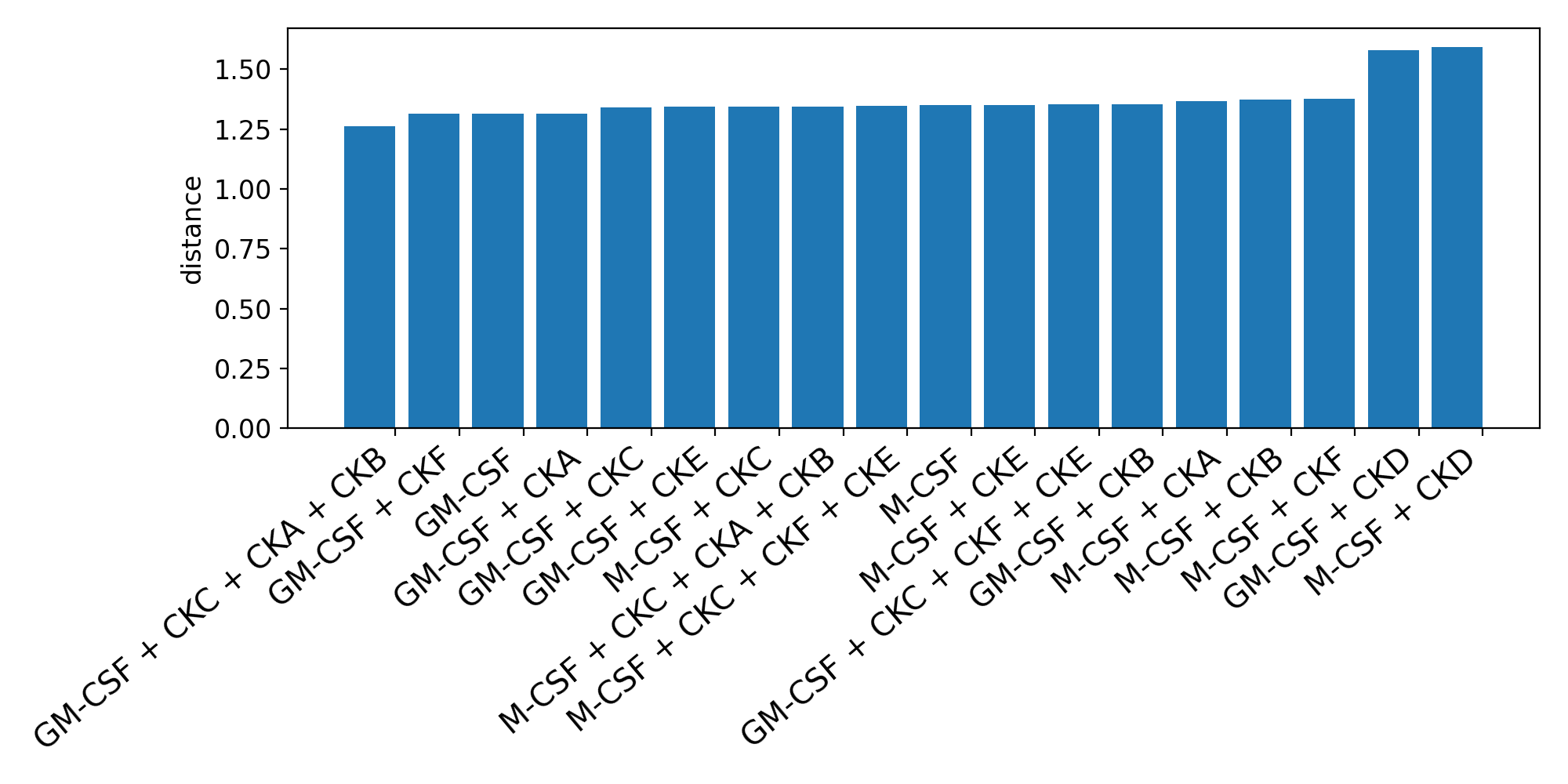}
    \caption{SystemMatch-produced ranking of \textit{in vitro} conditions based on their computed similarity to the target condition.}
    \label{fig:ranking}
\end{figure}

Fig.~\ref{fig:dataset}a shows the UMAP~\citep{umap} dimensionality reduction of the integrated macrophage atlas annotated by its marker genes. We identify inflammatory, suppressive, proliferating, and co-suppressive/inflammatory cell subsets, the last of which is not typically described in reviews of macrophage biology. For the purposes of this study, we target the suppressive population for \textit{in vitro} system optimization.

To study the heterogeneity of macrophage development systems, we differentiated monocytes \textit{in vitro} with either \textit{M-CSF} or \textit{GM-CSF} for 6 days and then stimulated these macrophages for an additional 4 days with one of nine combinations of cytokines typically understood to play a role in regulatory macrophage phenotype. Fig.~\ref{fig:dataset}b shows the UMAP dimensionality reduction of the \textit{in vitro} systems. We see that these systems primarily separate into three distinct groups, which can be described as pro-inflammatory, pro-suppressive, and basal.

Finally, to compare the \textit{in vitro} cells to our human tumor macrophage atlas, we show a heatmap of selected marker genes for both datasets (Fig.~\ref{fig:dataset}c). We observe qualitatively that not one of the \textit{in vitro} systems fully recapitulates the gene expression of the suppressive \textit{gene36-gene46+} population. This is further recapitulated by the similarity ranking of the \textit{in vitro} systems to the target population (Fig.~\ref{fig:ranking}), in which a) the best-ranked system using M-CSF, which serves as the basis for all systems most commonly used to generate suppressive macrophages \citep{mia2014optimized,fogg2020ovarian} is ranked 7th; and b) the closest system only reduces the distance to the target by 20\% from the least common system, which is well known to produce a pro-inflammatory response.

\begin{figure}
     \centering
     \begin{subfigure}[t]{0.3\textwidth}
         \vskip 0pt
         \centering
         \includegraphics[width=\textwidth]{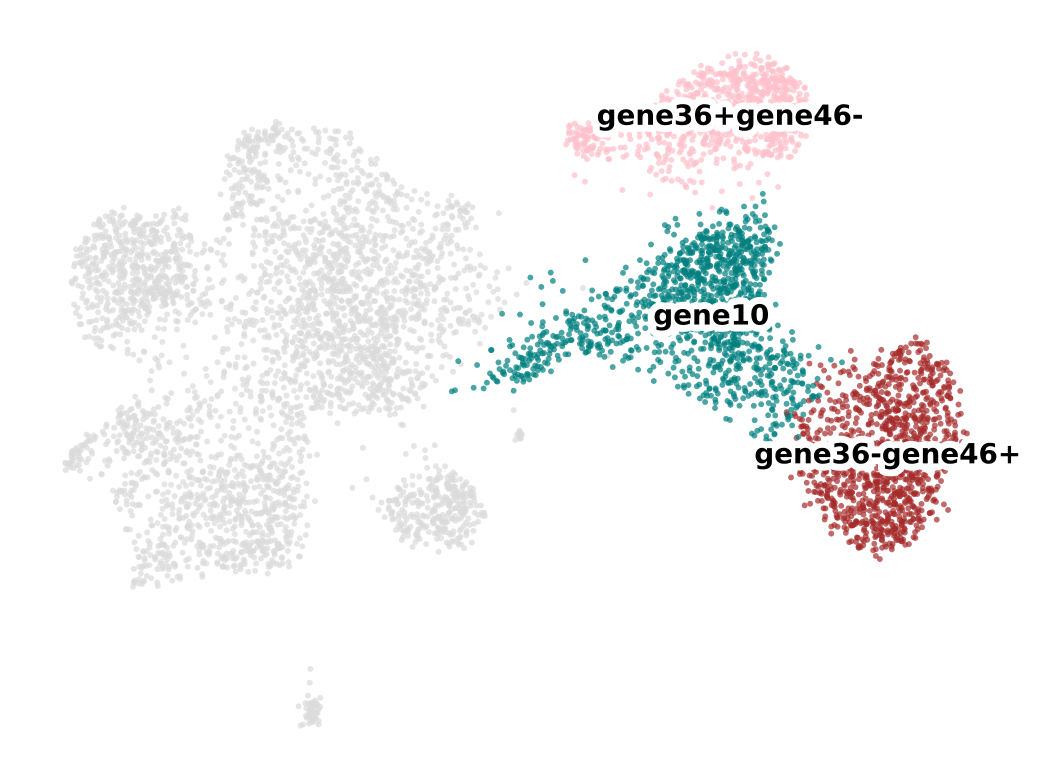}
     \end{subfigure}
     \hfill
     \begin{subfigure}[t]{0.34\textwidth}
         \vskip 0pt
         \centering
         \includegraphics[width=\textwidth]{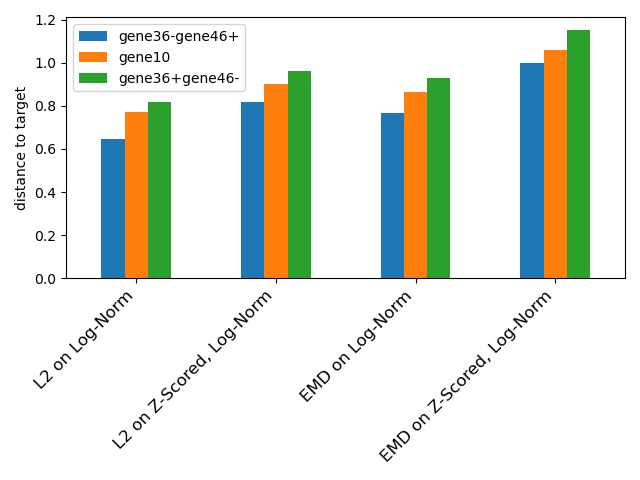}
     \end{subfigure}
     \hfill
     \begin{subfigure}[t]{0.34\textwidth}
         \vskip 0pt
         \centering
         \includegraphics[width=\textwidth]{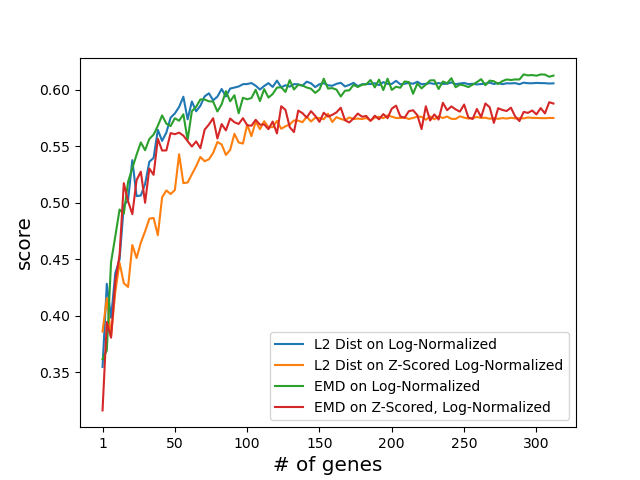}
     \end{subfigure}
     \hfill
    \caption{Quantitative evaluation of distance metrics for ranking data with known ground truth. UMAP (left) shows the annotated held-out, ground truth macrophage data; computed distances (middle) between the held-out macrophage populations and the target population for each method; evaluation scores (right) of each distance metric relative to the known ground truth.}
    \label{fig:evaluation}
\end{figure}

\subsection{Evaluating similarity ranking methods}

To validate the effectiveness of SystemMatch, we compare held-out ground truth macrophage data from \citet{tang2018human} to our human tumor macrophage atlas. We preprocessed the held-out data by filtering out low-quality cells and annotated cells in the dataset using the same gene set and protocol as for the atlas, marking the macrophages as highly suppressive, moderately suppressive, or nonsuppressive. Fig.\ref{fig:evaluation} (left) shows the UMAP of the final annotated clusters of the ground truth macrophage dataset. 

Because our target cells are suppressive macrophages, we expected the ranking of the ground truth cells regarding their proximity to the target suppressive cells to be from most suppressive (\textit{gene36-gene46+}) to least suppressive (\textit{gene36+gene46-}). The following distance metrics were considered (see Methods): 1) L2 distance on log-normalized pseudobulked gene expression; 2) L2 distance on z-scored, log-normalized pseudobulked gene expression; 3) Earth Mover's Distance on log-normalized single cell gene expression; 4) Earth Mover's Distance on z-scored, log-normalized single cell gene expression. On the full dataset, all four distance metrics correctly ranked the three datasets (Fig.~\ref{fig:evaluation} (middle)), so we define the following score to quantitatively evaluate a) how well it correctly ranked the held-out conditions and b) the separation of the most suppressive condition from the least suppressive condition.

For a given distance metric $m$, query subsets $\{\sX_i ~|~ i \in \{\textit{gene36-gene46+}, \textit{gene10}, \textit{gene36+gene46-}\}$ and target population $\sY$, we compute distances between the conditions in the held-out dataset to the target condition giving a distance vector 
$$\evd^m = \left( m(\sX_{\textit{gene36-gene46+}}, \sY),  m(\sX_{\textit{gene10}}, \sY),  m(\sX_{\textit{gene36+gene46-}}, \sY) \right)$$
with corresponding rank vector $\evr^m$, where the expected rank vector $\evr^{
\textit{true}} = (1, 2, 3)$. For a ranking $r_m$ of length $n$, we define the score

$$\mathrm{Score(m)} = \left( \frac{\sum_i^n I(\evr^m_i = r^{\textit{true}}_i)}{n} + \frac{\evd^m_{gene36+gene46-} - \evd^m_{gene36-gene46+}}{\evd^m_{gene36+gene46-}}\right) / 2$$

We compute this score for different levels of dataset corruption, which we implement by subsampling genes to remove information. We repeat this subsampling up to 200 times, or until the score converges. Fig.~\ref{fig:evaluation} (right) shows the scores for the four metrics; EMD with log-normalized genes performs best with the full gene set, but pseudo-bulked L2 distance is most robust to corruption.

\subsection{Recommending future experiments}

\begin{figure}[tb]
\begin{center}

     \centering
     \begin{subfigure}[b]{0.30\textwidth}
         \centering
         \includegraphics[width=\textwidth]{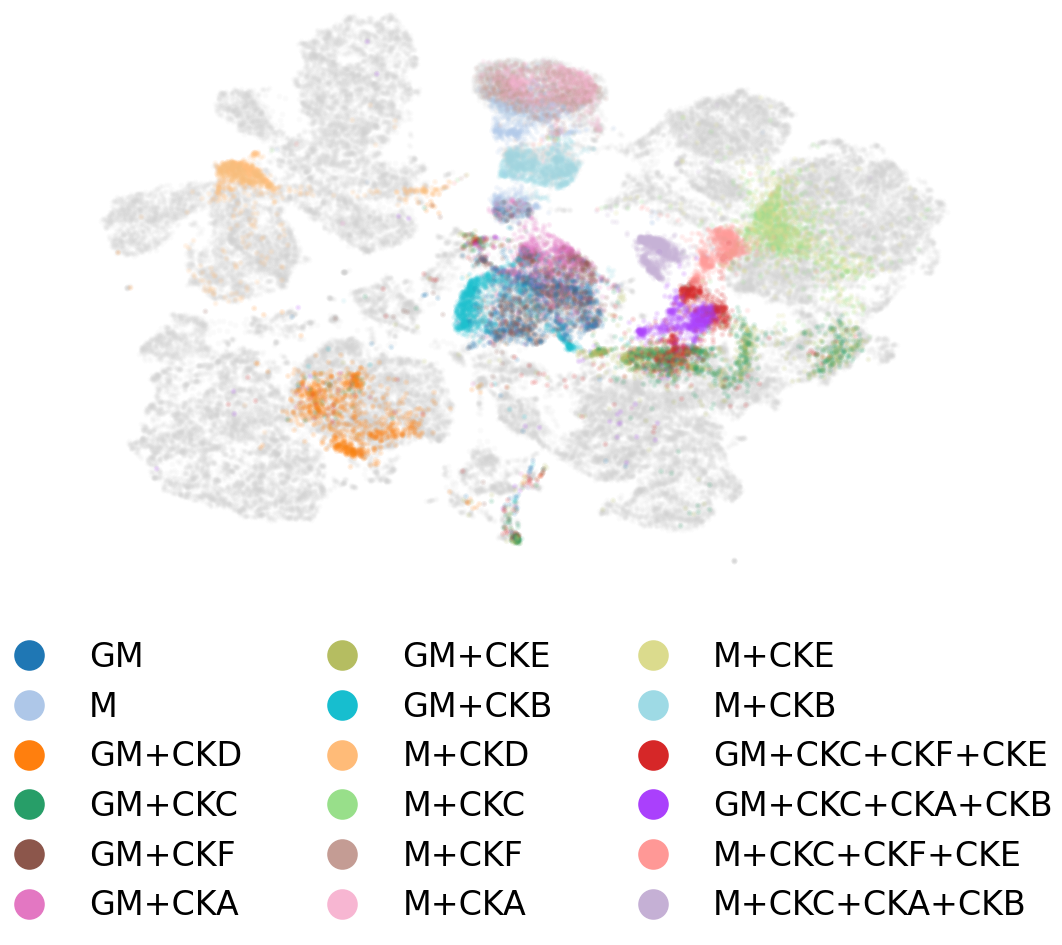}
     \end{subfigure}
     \hfill
     \begin{subfigure}[b]{0.36\textwidth}
         \centering
         \includegraphics[width=\textwidth]{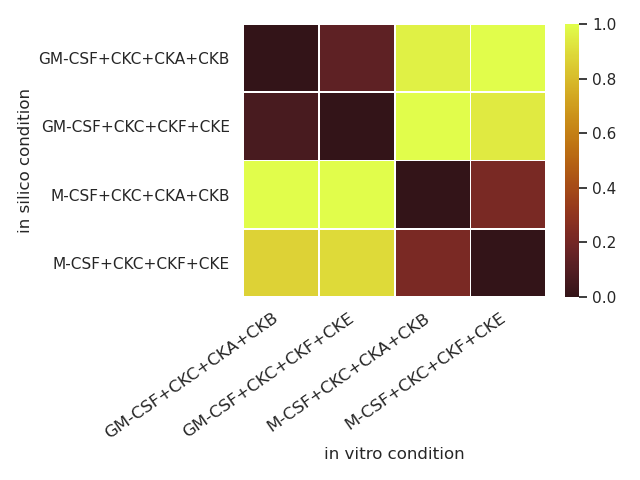}
     \end{subfigure}
     \hfill
     \begin{subfigure}[b]{0.25\textwidth}
         \centering
         \includegraphics[width=\textwidth]{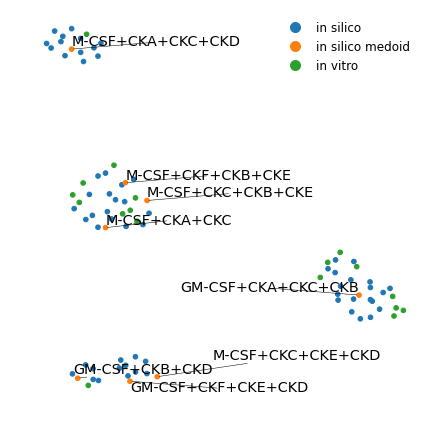}
     \end{subfigure}
\end{center}
\caption{Sample output of the experiment recommender system. UMAP embedding of \textit{in vitro} and \textit{in silico} single cells (left), with \textit{in silico} predictions shown in grey; heatmap of normalized distances from predicted triplets to held-out triplets (middle); UMAP embedding of average expression per condition (right), with $k$-medoids--recommended perturbations highlighted in orange.}
\label{fig:recommender}
\end{figure}

In order to leverage the recommendations of SystemMatch to further improve the \textit{in vitro} systems beyond those initially tested, we combine compositional perturbational autoencoders with a modified $k$-medoids algorithm to select a subset of possible combinations of the initially tested cytokines to test in the next experimental iterations. First, we use a modified form of CPA~\citep{lotfollahi2021compositional} to generate all possible double- and triple-combinations of cytokines (Fig.~\ref{fig:recommender} (left)). We see that \textit{M/GM-CSF} and \textit{CKD} drive the most significant differences in the predicted cells, which is consistent with our understanding of these cytokines. Further, we see as expected that the double-combinations of cytokines are generally more similar to the \textit{in vitro} single cytokine perturbations. 

To validate the accuracy of our \textit{in silico} predictions, we compared the similarity of the held-out \textit{in vitro} triplet perturbations to the corresponding \textit{in silico} perturbations. Fig.~\ref{fig:recommender} (middle) shows the distance (using the L2 pseudobulk distance normalized to a range of $[0, 1]$) from each of the held-out triplets to the corresponding predicted triplets. All of the four are closest to the corresponding prediction. Equivalent heatmaps for the other distance metrics are shown in Fig.~\ref{fig:triplets}.

Next, we run our modified $k$-medoids algorithm on the average expression on each condition to select a subset of these double- and triple-combinations to test experimentally in future work to maximize the heterogeneity of all tested perturbations. Alternatively, we can use SystemMatch to test the proximity of the generated conditions; this will produce a less heterogeneous set of combinations for further testing but will optimize more directly towards the target condition. The result of this comparison is shown in Tab.~\ref{fig:ranking_in_silico}; interestingly, the top predicted condition uses M-CSF, even though none of the top six \textit{in vitro} conditions did. The \textit{in silico} preference (however mild) for \textit{M-CSF} over \textit{GM-CSF} is consistent with commonly used models of suppressive macrophages \citep{mia2014optimized,fogg2020ovarian}, giving confidence in our recommendations.

\section{Discussion}

In this work we provide an end-to-end ML pipeline for assessing the fitness of and optimizing preclinical models to maximize their predictive power of clinical results. We do this through an iterative process of comparing model systems to the target \textit{in sapiens} population, applying perturbation prediction to suggest experimental changes, and iterating upon this process to produce preclinical models that are more representative of the relevant tumor or tissue data. We demonstrate our pipeline through a use case of developing an \textit{in vitro} system optimized to produce macrophages most similar to human tumor-derived suppressive macrophages, and we recommend a set of further experiments to fully explore the space of possible perturbations and ultimately generate a system more suited to developing drugs targeting these cells than those systems currently used in the literature.

We validate our pipeline's performance through a series of tests using ground truth data with a held-out \textit{in sapiens} public dataset in which we manually annotate cells known to be close and far from the target population. We show that a) our pipeline is largely robust to the choice of distance metric; b) distance metrics computed on average expression are generally most robust to data corruption; and c) in the limit of low data corruption, single-cell optimal transport metrics may outperform these "pseudobulk" methods. We then validate our perturbation prediction using experimentally generated \textit{in vitro} controls of cytokine combinations, through which we show that the \textit{in silico} generated combinations are most similar to the corresponding matching \textit{in vitro} populations.

We note that the robustness to the use of average expression (compared to single-cell) indicates that bulk RNA-seq could suffice in some cases for the query datasets; however, this would preclude the use of CPA, and the resulting lower-resolution data would provide less predictive power to generate \textit{in silico} samples. Additionally, in the case of heterogeneous model systems (e.g. in systems with developmental trajectories or large numbers of cycling cells), single-cell metrics may score more highly. We also note that the use of \textit{in silico} generated combinations of experimental conditions explores only a subset of possible model systems; systems generated with conditions not included in the initial experiment cannot be discovered in this way, and must instead be added to the iterative process by manual expert review.

In recent years, the utility of preclinical model data to evaluate the clinical relevance of a drug has been questioned. We show here, through a combination of single-cell genomic data and machine learning, a method by which these preclinical model systems can be optimized to enhance their potential predictive power. While ML methods have generated a great deal of impact in target discovery and validation, this is to our knowledge the first ML pipeline for actually establishing the preclinical model systems in which those targets can be evaluated.



\bibliography{main}
\bibliographystyle{iclr2022_conference}

\appendix
\section{Appendix}

\subsection{CPA Model Training Details}

The CPA model was trained, tested, and evaluated on the \textit{in vitro} dataset, in which cells were treated in eighteen separate conditions. Of these conditions, two were designated as control conditions (\textit{GM-CSF} and \textit{M-CSF}), four were considered triplet combinations of perturbations (e.g., \textit{GM-CSF} + \textit{CKC} + \textit{CKA} + \textit{CKB}), and the other twelve were single cytokine perturbations (e.g., \textit{GM-CSF} + \textit{CKF}). Since CPA is able to learn an independent transcriptional embedding per single perturbation and then add these embeddings to predict multiplet perturbations, we chose to train and validate the model on the control and singular cytokine perturbation conditions and evaluate its performance on the triplet perturbation conditions. The training dataset consisted of 417 genes deemed relevant to macrophage function by a combination of domain expertise and data-driven highly variable genes \citep{yip2018hvg} from our human tumor macrophage atlas.

\begin{figure}[htbp]
    \centering
    \begin{subfigure}[t]{0.49\textwidth}
        \centering
        \includegraphics[width=\textwidth]{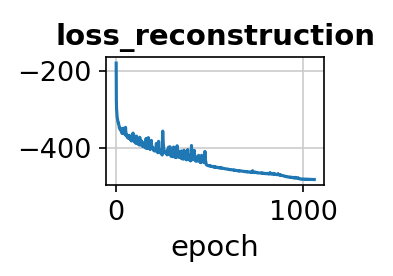} \\
        (a)
    \end{subfigure}
    \hfill
    \begin{subfigure}[t]{0.49\textwidth}
        \centering
        \includegraphics[width=\textwidth]{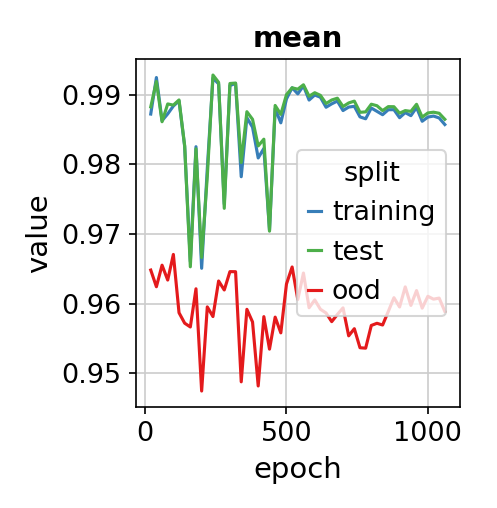} \\
        (b)
    \end{subfigure}
    \caption{Evaluation of model performance. (a) Reconstruction loss per training epoch of the CPA model. (b) R-squared between mean predicted gene expression versus actual gene expression over all 417 genes.}
    \label{fig:cpa_performance}
\end{figure}

We ran CPA with an autoencoder width of 512, batch size of 128, and embedding size of 128, and we trained it for six hours over 1060 epochs on an NVIDIA Tesla A100 GPU. The reconstruction loss and model performance on the held-out triplet conditions (OOD) are shown in Fig.~\ref{fig:cpa_performance}. We used this model to predict additional conditions, all of which were doublet or triplet cytokine conditions that were not present in the \textit{in vitro} assay, in order to investigate which would best match our target model system.  

\subsection{Generation of \protect\textit{in vitro} macrophage model systems}

Human CD14+ Monocytes were isolated from peripheral mononuclear blood cells (PBMCs) using the EasySep™ Human Monocyte Isolation Kit (STEMCELL Technologies) following manufacturer’s instructions.  Monocytes were plated at 1x10\textsuperscript{6} cells per well in 6 well tissue culture treated plates (Corning).  Cells were plated in 2.5mL RPMI media (Life Technologies) supplemented with 10\% FBS (Life Technologies)  and 1\% Penicillin-Streptomycin (Thermo Fisher Scientific) and either \textit{M-CSF} (PeproTech) or \textit{GM-CSF} (PeproTech), respectively, at 20ng/mL.  At day 3, media was aspirated and 2.5mL fresh \textit{M-CSF/GM-CSF} media was added.  At day 6, media was aspirated and 3mL \textit{M-CSF/GM-CSF} media supplemented with additional differentiation cytokines was added to respective wells.  Cells were isolated at day 10 with cell scrapers (Fisher Scientific) and counted with the Cellaca MX High-throughput Automated Cell Counter (Nexcelom) following manufacturer’s instructions prior to single cell processing.

\begin{figure}[htbp]
    \centering
    \begin{subfigure}[t]{0.49\textwidth}
        \centering
        \includegraphics[width=\textwidth]{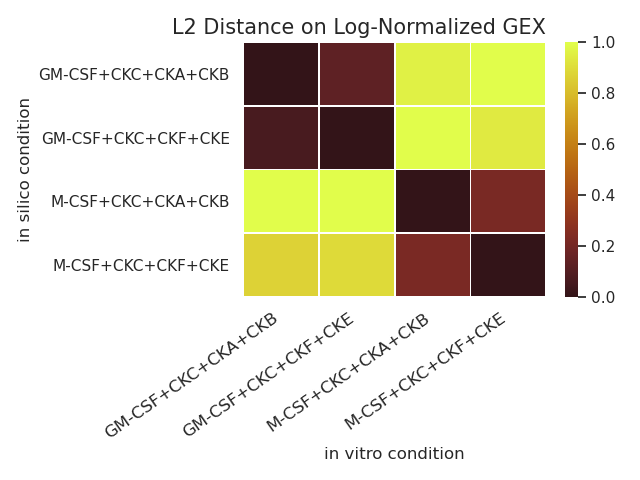} \\
        (a) \\
        \includegraphics[width=\textwidth]{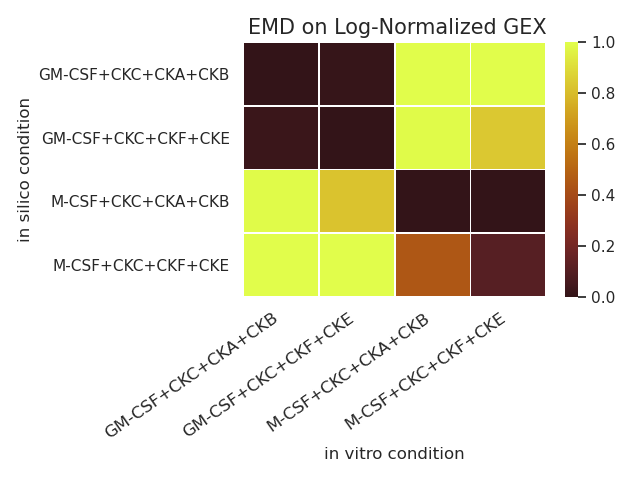} \\
        (c)
    \end{subfigure}
    \hfill
    \begin{subfigure}[t]{0.49\textwidth}
        \centering
        \includegraphics[width=\textwidth]{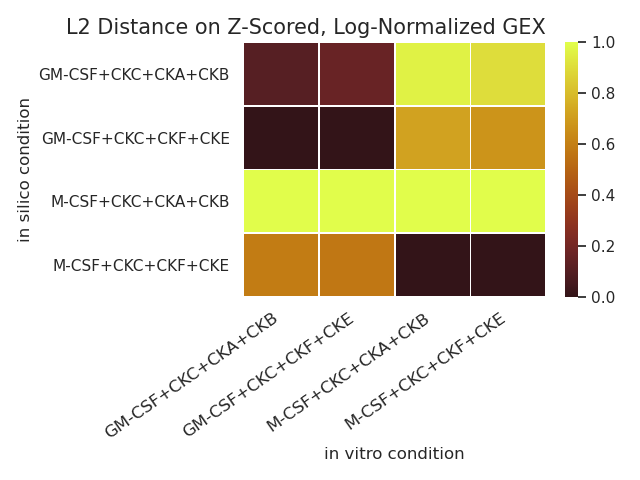} \\
        (b) \\
        \includegraphics[width=\textwidth]{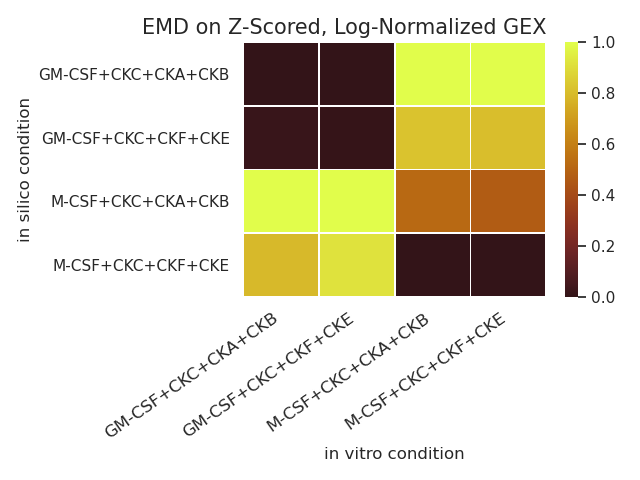} \\
        (d)
    \end{subfigure}
    \caption{Evaluation of held-out \textit{in vitro} triplet controls' proximity to predicted \textit{in silico} triplets using all four tested distance metrics. Distances are normalized by column to the range $[0,1]$. (a) L2 distance on log-normalized; (b) L2 distance on z-scored log-normalized; (c) EMD on log-normalized; (d) EMD on z-scored log-normalized.}
    \label{fig:triplets}
\end{figure}

\begin{figure}[htbp]
\centering
\includegraphics[width=\textwidth]{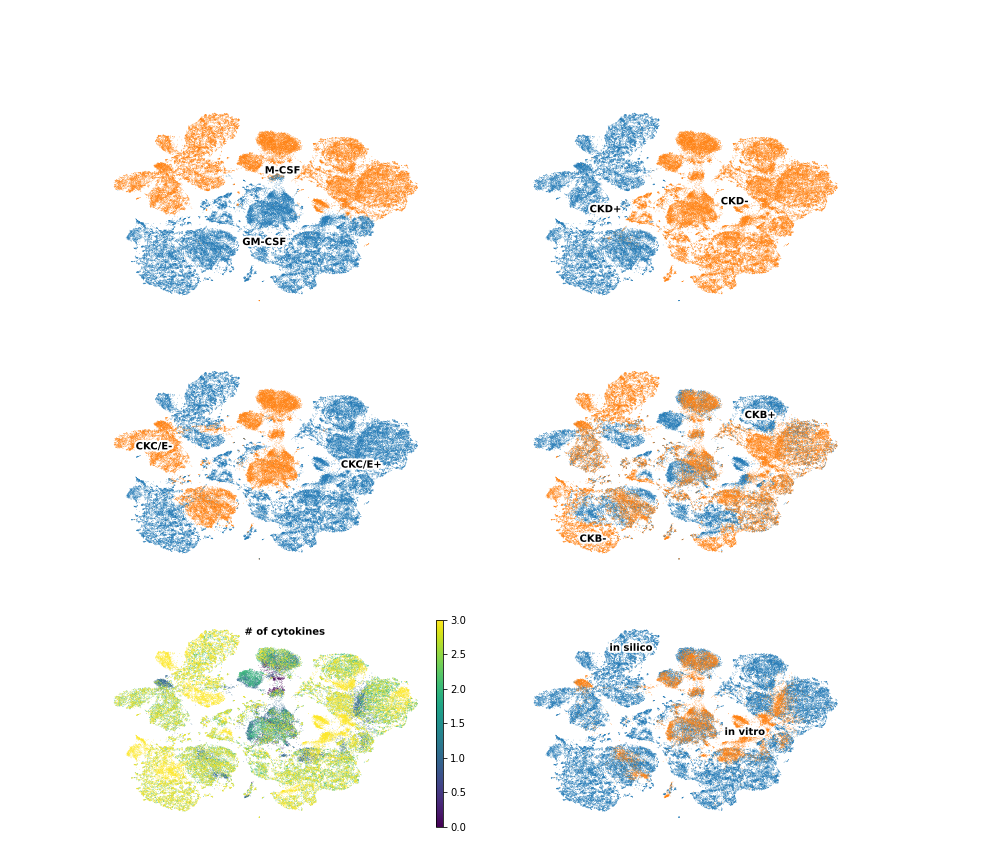}
\caption{Detailed characterization of \textit{in silico} and \textit{in vitro} combined samples.}
\label{fig:umap_in_silico}
\end{figure}

\begin{figure}[htbp]
\centering
\includegraphics[width=\textwidth]{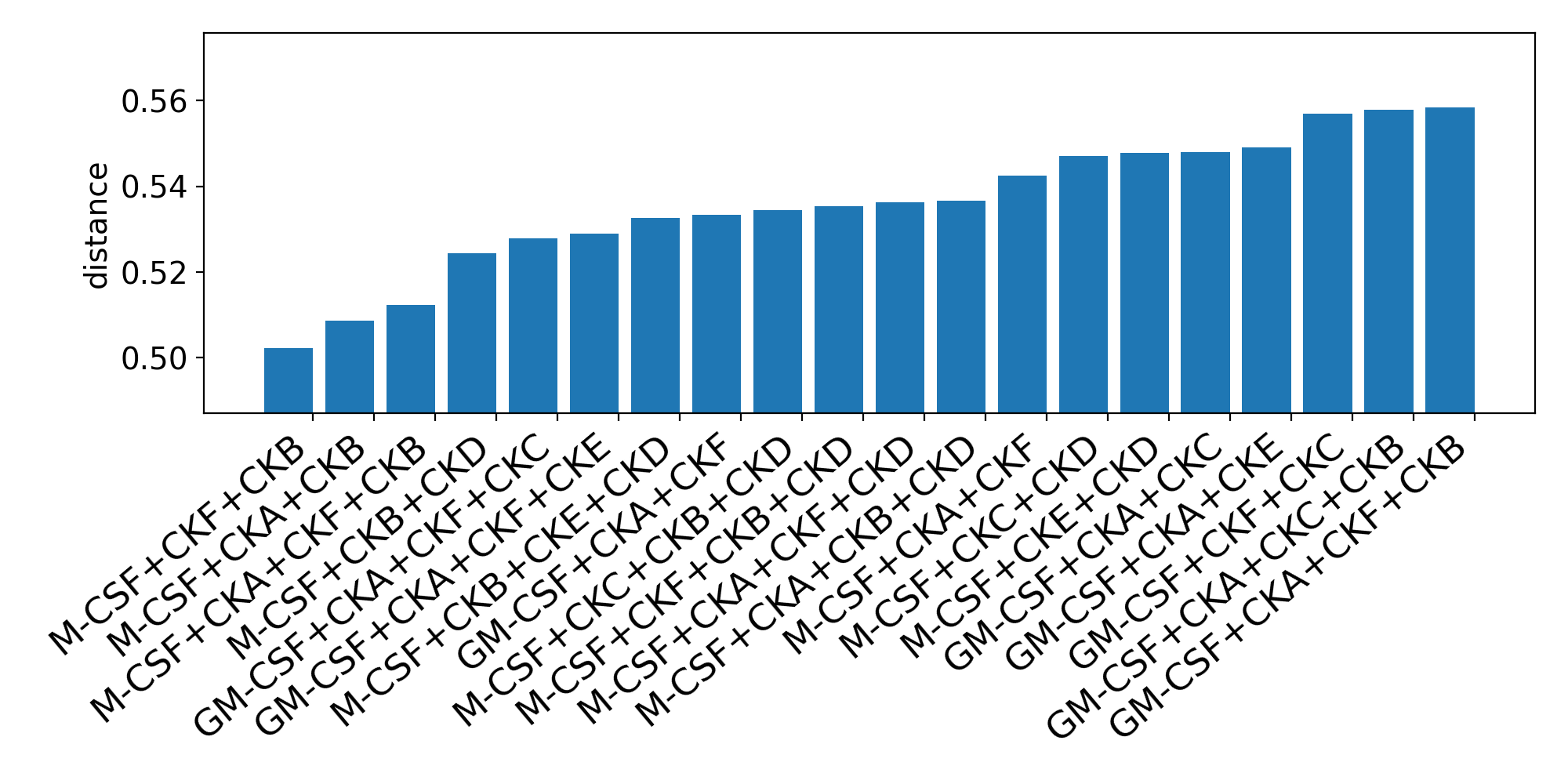}
\caption{SystemMatch-produced ranking of top 20 \textit{in silico} conditions based on their computed similarity to the target condition.}
\label{fig:ranking_in_silico}
\end{figure}

\end{document}